\documentclass[letterpaper]{article} 
\usepackage{aaai25}  
\usepackage{times}  
\usepackage{helvet}  
\usepackage{courier}  
\usepackage[hyphens]{url}  
\usepackage{graphicx} 
\usepackage{natbib}  
\usepackage{caption} 
\usepackage{algorithm}
\usepackage{algorithmic}

\usepackage{newfloat}
\usepackage{listings}
\DeclareCaptionStyle{ruled}{labelfont=normalfont,labelsep=colon,strut=off} 
\floatstyle{ruled}
\newfloat{listing}{tb}{lst}{}
\floatname{listing}{Listing}

\usepackage{amsthm}
\newtheorem{theorem}{Theorem}[section]
\theoremstyle{definition}
\newtheorem{definition}{Definition}
\title{Higher Order Structures For Graph Explanations}
\author{
    Akshit Sinha\equalcontrib\textsuperscript{\rm 1},
    Sreeram Vennam\equalcontrib\textsuperscript{\rm 1},
    Charu Sharma\textsuperscript{\rm 1},
    Ponnurangam Kumaraguru\textsuperscript{\rm 1}
}
\affiliations{
    \textsuperscript{\rm 1}International Institute of Information Technology, Hyderabad\\
    \{akshit.sinha, sreeram.vennam\}@students.iiit.ac.in

}

\usepackage[acronym]{glossaries-extra}
\setabbreviationstyle[acronym]{long-short}

\newacronym{gnn}{GNN}{Graph Neural Network}
\newacronym{forge}{FORGE}{\textbf{F}ramework For Higher-\textbf{O}rder \textbf{R}epresentations In \textbf{G}raph \textbf{E}xplanations}

\usepackage{booktabs}
\usepackage{multirow}
\usepackage{multicol}

\usepackage{xcolor}

\begin{document}

\maketitle

\begin{abstract}

Graph Neural Networks (GNNs) have emerged as powerful tools for learning representations of graph-structured data, demonstrating remarkable performance across various tasks. Recognising their importance, there has been extensive research focused on explaining GNN predictions, aiming to enhance their interpretability and trustworthiness. However, GNNs and their explainers face a notable challenge: graphs are primarily designed to model pair-wise relationships between nodes, which can make it tough to capture higher-order, multi-node interactions. This characteristic can pose difficulties for existing explainers in fully representing multi-node relationships. To address this gap, we present \textbf{F}ramework For Higher-\textbf{O}rder \textbf{R}epresentations In \textbf{G}raph \textbf{E}xplanations (FORGE), a framework that enables graph explainers to capture such interactions by incorporating higher-order structures, resulting in more accurate and faithful explanations. Extensive evaluation shows that on average real-world datasets from the GraphXAI benchmark and synthetic datasets across various graph explainers, FORGE improves average explanation accuracy by 1.9x and 2.25x, respectively. We perform ablation studies to confirm the importance of higher-order relations in improving explanations, while our scalability analysis demonstrates FORGE's efficacy on large graphs.

\end{abstract}

\section{Introduction}
\glspl{gnn} \cite{gnn} have become increasingly important in graph representation learning, as data in many real-world domains can be naturally modeled as graphs. \glspl{gnn} have found applications in several sensitive fields, including information processing \cite{ying_he_chen_eksombatchai_hamilton_leskovec_2018, wang_he_cao_liu_chua_2019}, criminal justice \cite{agarwal2021towards}, molecular chemistry \cite{gilmer_schoenholz_riley_vinyals_dahl_2017, duvenaud_maclaurin_aguilera}, and bio-informatics \cite{10.3389/fgene.2021.690049, fout_byrd_shariat_ben-hur_2017, mfse}.
In these sensitive domains, interpretability is crucial for ensuring transparent, justifiable, and ethical decision-making. As GNN usage expands, understanding their internal processes becomes vital for effective and safe usage of these models in practical settings.
To address these challenges, various graph explainers have been proposed.
These graph explainers typically provide two types of explanations crucial for \gls{gnn} prediction: (1) identification of subgraphs \cite{subgraphx, graph_mask}, and (2) determination of node features \cite{graphlime, gnnexplainer}.

\begin{figure}[t]
\centering
\includegraphics[width=7cm]{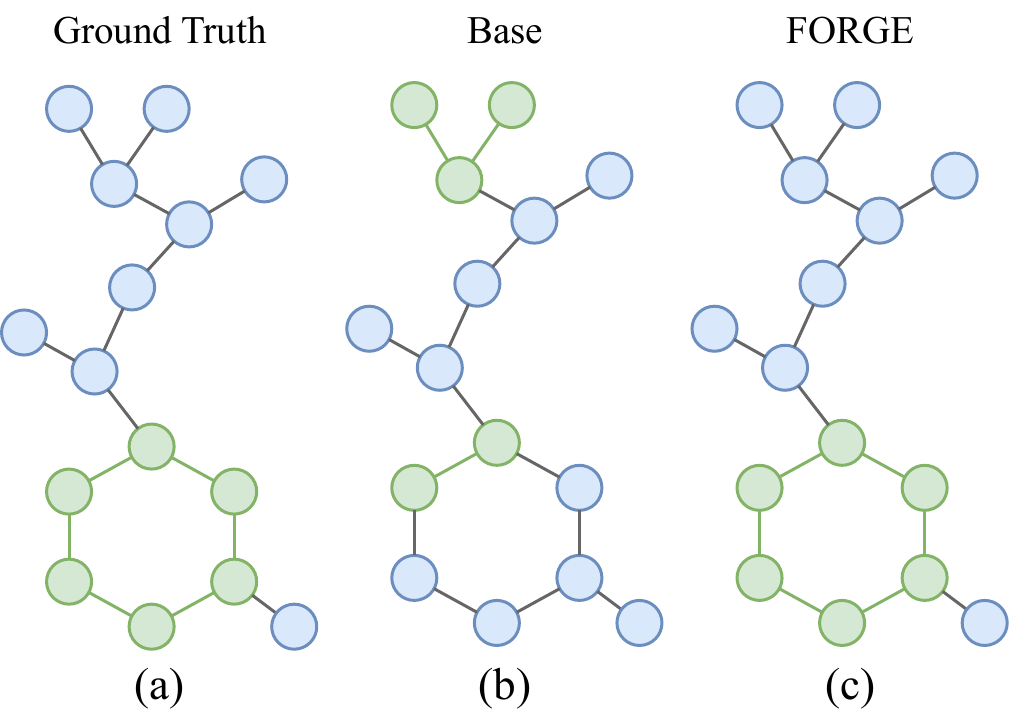}
\caption{(a) Ground Truth for an example from BENZENE \cite{graphxai} (b) Explanation generated by GNNExplainer \cite{gnnexplainer} (c) Explanation generated by using FORGE. \textbf{Green} nodes and edges signify the subgraph considered important for GNN prediction. By incorporating FORGE, we can capture important multi-node interactions, resulting in more accurate explanations.}
\label{fig:front_page}
\end{figure}

However, graph explainers and \glspl{gnn} face an inherent limitation in their ability to generate representations stemming from graphs' ability to model only pair-wise interactions between nodes. This limitation has implications on the expressivity of \glspl{gnn}, which is explored in detail by \citet{gin}. To address this, higher-order structures such as \textit{cell complexes} \cite{cellcomplex} have been employed as they are a higher-order generalization of graphs, capable of modeling multi-node interactions. Neural networks designed for these higher-order structures \cite{cwn, snn, efficientsnn} have demonstrated significant performance improvements in graph learning tasks compared to traditional graph structures, and have been shown to be more expressive than traditional \glspl{gnn}.
Despite these advancements in higher-order graph representations, there exists a critical gap in the field of graph explainability: the potential of higher-order structures to enhance the interpretability of graph-based models remains unexplored.

In this work, we present \gls{forge}, a novel, explainer-agnostic framework designed to enhance the capability of graph explainers by capturing multi-node interactions during the \gls{gnn} learning process itself by internally representing the underlying graph data as a cell complex. To further refine the quality of explanations, as part of our framework we introduce Information Propagation algorithms that translate the explanations generated for cell complexes back to the underlying graph data, resulting in richer, more accurate, and faithful explanations with respect to the input data as well as the underlying GNN predictor. This is encapsulated by Figure \ref{fig:front_page}, where our framework can capture the multi-node interactions of a benzene ring, which is the correct ground truth explanation. Without FORGE, the base explainer is unable to interpret these interactions, resulting in less accurate explanations.

\begin{figure}[t]
\centering
\includegraphics[width=7cm]{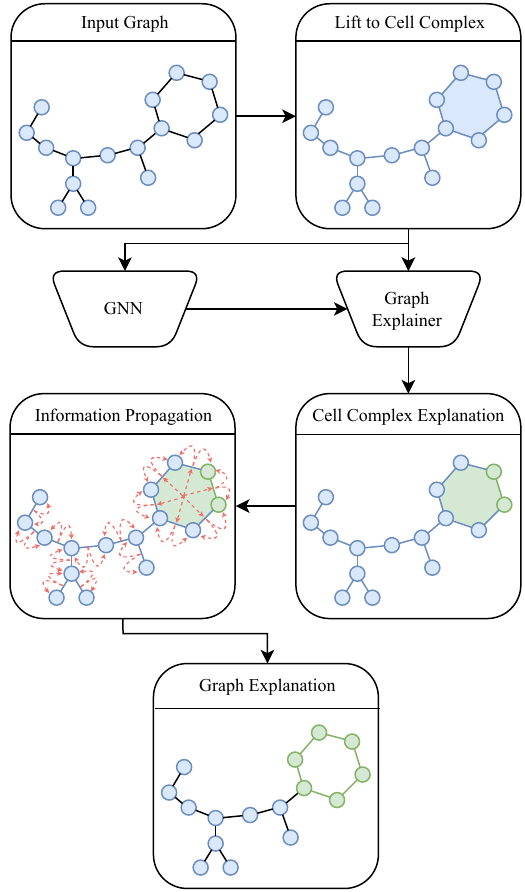}
\caption{Visual representation of FORGE. The input graph is lifted to a cell complex, which is then given as input to (i) a GNN to train on, as well as to (ii) a graph explainer. Propagation is then done on the output cell complex explanation to map it to an explanation for the original graph. The green color on cells, nodes, and edges signify the substructure considered important for GNN prediction (the explanation).}
\label{fig:framework}
\end{figure}

The overall framework is described in Figure \ref{fig:framework}. We conduct extensive evaluations of \gls{forge} on real-world datasets from the GraphXAI \cite{graphxai} benchmark, as well as on specially curated synthetic datasets. Our results demonstrate that incorporating \gls{forge} consistently matches or improves the explanation accuracy and faithfulness of various graph explainers. To reaffirm our hypothesis, we perform rigorous ablations and find that each components of our framework contribute significantly to increased explainer performance. Additionally, our analysis on scalability reveals that \gls{forge} efficiently handles dense, complex graph networks with only linear overhead in both time and space complexity, further emphasizing its practical applicability and computational efficiency.

\section{Related Work}

\subsubsection{Higher-Order Representations of Graphs}

\citet{cellcomplex} first explored relating higher-order structures to spectral graph theory. Following this, recent advancements in \gls{gnn} architecture have explored the incorporation of higher-order topological structures, such as simplicial complexes \cite{snn} and cell complexes \cite{cwn}, to successfully enhance the expressive power and representational capacity of these models. This line of work is motivated by the inherent limitations of traditional GNNs, particularly their constraints in capturing complex structural patterns \cite{gin} and effectively modeling long-range dependencies \cite{longrange}.

\subsubsection{Explainability in \glspl{gnn}}

The increasing adoption of \glspl{gnn} in critical domains has resulted in significant research into methods for explaining their predictions. A comprehensive survey by \citet{survey} provides an extensive overview of \gls{gnn} explainability techniques.

Perturbation-based methods, such as GNNExplainer \cite{gnnexplainer}, PGExplainer \cite{pgexplainer}, SubgraphX \cite{subgraphx}, and GraphMask \cite{graph_mask}, identify important subgraphs by systematically perturbing input graphs and analyzing the impact on model outputs. Gradient-based approaches, including Grad-CAM \cite{gradcam} and Guided Backpropagation \cite{guidedbp}, leverage gradient information to assess the influence of input features on model predictions. Surrogate methods, exemplified by PGM-Explainer \cite{pgmexplainer} and GraphLIME \cite{graphlime}, approximate complex GNN behavior with interpretable local models.
Recent innovations have further expanded the landscape of GNN explainability. RG-Explainer \cite{rlexplainer} employs reinforcement learning to construct explanatory subgraphs tailored to both model architecture and individual instances. MATE \cite{mate} introduces a meta-learning framework that jointly optimizes GNN performance and explainability, resulting in intrinsically more interpretable representations. MotifExplainer \cite{motifexplainer} introduces a motif-based approach for human interpretable explanations. While MotifExplainer is most similar to our work, it is domain specific and suffers from $O(n^3)$ complexity.

Although these approaches have significantly advanced GNN explainability, they primarily operate on traditional graph structures. Our proposed framework, \gls{forge}, distinguishes itself by leveraging higher-order topological structures, specifically cell complexes, (1) to train GNNs to inherently create more interpretable representations during the learning process, and (2) utilise their topological properties to refine explanations post-hoc.

\section{Cell Complexes}\label{sec:complex}
Graphs are powerful structures that excel in modelling relations between objects.
However, they are limited to modelling pairwise relationships.
To overcome this, graphs can be generalised to work in higher dimensions and model group-wise relationships.
This generalisation in algebraic topology is achieved by \textit{cell complexes} \cite{cellcomplex}.

A cell complex $X$ is constructed through a hierarchical assembly process.
It begins with a set of vertices (0-cells), to which edges (1-cells) are attached by connecting these points with closed line segments, forming a graph.
This structure is then expanded by attaching faces (2-cells) to corresponding 1-cells.
While our focus remains within two dimensions, this process can be extended to higher dimensions\footnote{For a more comprehensive understanding of cell complexes, we point the reader to \url{https://www.math.ksu.edu/~hansen/CWcomplexes.pdf} and \citet{cwn}}.

While our work does not go into detail about the algebraic topology ideas linked to \textit{cell complexes}, it is important to provide some basic definitions \cite{cellcomplex} and notations (Table \ref{tab:notations}) that are essential to understand the framework we are proposing.

\begin{table}[h]
    \centering
    \begin{tabular}{cc}
        \toprule
        \textbf{Notation} & \textbf{Description}  \\ \midrule
        $G$ & A general graph \\
        $V$ & Set of all vertices in a graph \\
        $E$ & Set of all edges in a graph \\
        $G(V,E)$ & A graph with vertices $V$ and edges $E$  \\ \midrule
        $X$ & A general cell complex \\
        $X^{(p)}$ & $p$-skeleton of a cell complex\\
        $c$, $c^{(p)}$ & general cell, cell of dimension $p$ \\
        $C^{(p)}$ & A $p$-chain in a given cell complex \\
        $C$ & Set of all $p$-chains in a cell complex \\
        $\sigma_{c_1, c_2}$ & A general boundary relation \\
        $\Sigma$ & Set of all boundary relations \\
        $X(C, \Sigma)$ & A cell complex defined by $C$ and $\Sigma$   \\ \bottomrule
        
    \end{tabular}
    \caption{Summary of notations used throughout the paper.}
    \label{tab:notations}
\end{table}

\begin{definition}[$p$-cell]
     A $p$-cell $c^{(p)}$ in a cell complex refers to an element of dimension $p$. In analogy to traditional graphs where we have vertices (0-dimensional) and edges (1-dimensional), cell complexes include these and extend to higher dimensions.
\end{definition}

\begin{definition}[\textit{boundary}/\textit{coboundary}]
     In a cell complex, a $p$-cell $c^{(p)}$ is considered a \textit{face} or \textit{boundary} of a $(p+1)$-cell $c^{(p+1)}$ if the set of points composing $c^{(p)}$ is a subset of those composing $c^{(p+1)}$. Conversely, $c^{(p+1)}$ is referred to as the \textit{coface} or \textit{coboundary} of $c^{(p)}$.
\end{definition}

\begin{definition}[$p$-chain]\label{def:pchain}
    In a given cell complex, a $p$-chain $C^{(p)}$ is simply defined as the set of all $p$-dimensional cells.
    The general set union of all such $C^{(p)}$ is denoted by $C$.
\end{definition}

\begin{definition}[$p$-skeleton]
     The $p$-skeleton of a cell complex $X$ is defined as the subcomplex $X^{(p)}$ consisting of cells of dimension at most $p$. Using this definition, we see that $X^{(0)}$ is the set of all vertices and $X^{(1)}$ is the set of all vertices and edges which precisely make up the underlying graph.
\end{definition}

\begin{definition}[\textit{boundary relation}]
     Within a cell complex, a boundary relation $\sigma$ is analogous to an edge in traditional graphs. It connects two cells, either of the same dimension (horizontal boundary relation) or different dimensions (vertical boundary relation). A horizontal boundary relation $\sigma_{c_1^{(p)}, c_2^{(p)}}$ links two $p$-cells that share a common \textit{boundary} or \textit{coboundary}, whereas vertical boundary relations $\sigma_{c_1^{(p)}, c_2^{(p+1)}}$ link a $p$-cell to its corresponding \textit{boundaries} or \textit{coboundaries}. In this work, we further restrict the boundary relations to be undirected, implying $\sigma_{c_1, c_2}$ = $\sigma_{c_2, c_1}$.
\end{definition}

Using the definitions provided in this section, we can analogously represent a cell complex as $X(C, \Sigma)$, like we represent a graph as $G(V, E)$. The methodology we introduce to create a cell complex from a given graph is described in Section \ref{sec:lift}.

While cell complexes are adept at handling higher-order interactions, their conventional definition requires them to be closed under taking subsets, making them inefficient data structures for scalable computation \cite{efficientsnn}. To overcome this, we present the following theorem and a subsequent modification in Equation \eqref{eq:theorem_eqn_3} to achieve an efficient, restricted form of cell complexes that is linear in space complexity. The proof follows by construction and is deferred to the Appendix.

\begin{theorem}\label{theorem}
For a graph $G(V, E)$ with adjacency matrix $A$ having cycles of length at most $K$, let $W_k$ represent the number of closed walks of length $k$ which are not $k$-cycles. Let $deg(v)$ represent the degree of a node $v$ in the graph. The corresponding cell complex $X$ will have cells $C$ and boundary relations $\Sigma$ such that

\begin{eqnarray}\label{eq:theorem_eqn_1}
\lvert C \rvert = |V| + |E| + \sum_{k=3}^K\frac{1}{2k}[tr({A^{(k)})} - W_k]
\end{eqnarray}
\begin{eqnarray}\label{eq:theorem_eqn_2}
\lvert \Sigma \lvert  \geq  3|E| + \frac{1}{2}\sum_{k=3}^K[tr({A^{(k)})} - W_k] + \sum_{v\in V}\binom{deg(v)}{2}
\end{eqnarray}

\end{theorem}

This theorem provides us with two important bounds. Equation \eqref{eq:theorem_eqn_1} shows that the number of cells in cell complex $X$ grows \textit{linearly} with the number of vertices and edges in the graph $G$. Equation \eqref{eq:theorem_eqn_2} shows that asymptotically, the size of $\Sigma$ grows \textit{quadratically} with $|E|$, due to the last term in the equation. This term is precisely the number of boundary relations that are present between two edges in the graph, which are represented by $\sigma_{c_1^{(1)}, c_2^{(1)}}$. To preserve the scalability and complexity of \gls{forge}, we further restrict how we create cell complexes from graphs, and drop all boundary relations of the form $\sigma_{c_1^{(p)}, c_2^{(p)}}$ where $p \geq 1$ from our construction of cell complexes. Doing this reduces Equation \eqref{eq:theorem_eqn_2} to:

\begin{eqnarray}\label{eq:theorem_eqn_3}
\lvert \Sigma \lvert  =  3|E| + \frac{1}{2}\sum_{k=3}^K[tr({A^{(k)})} - W_k]
\end{eqnarray}

\begin{figure}[t]
    \centering
    \includegraphics[width=8.43cm]{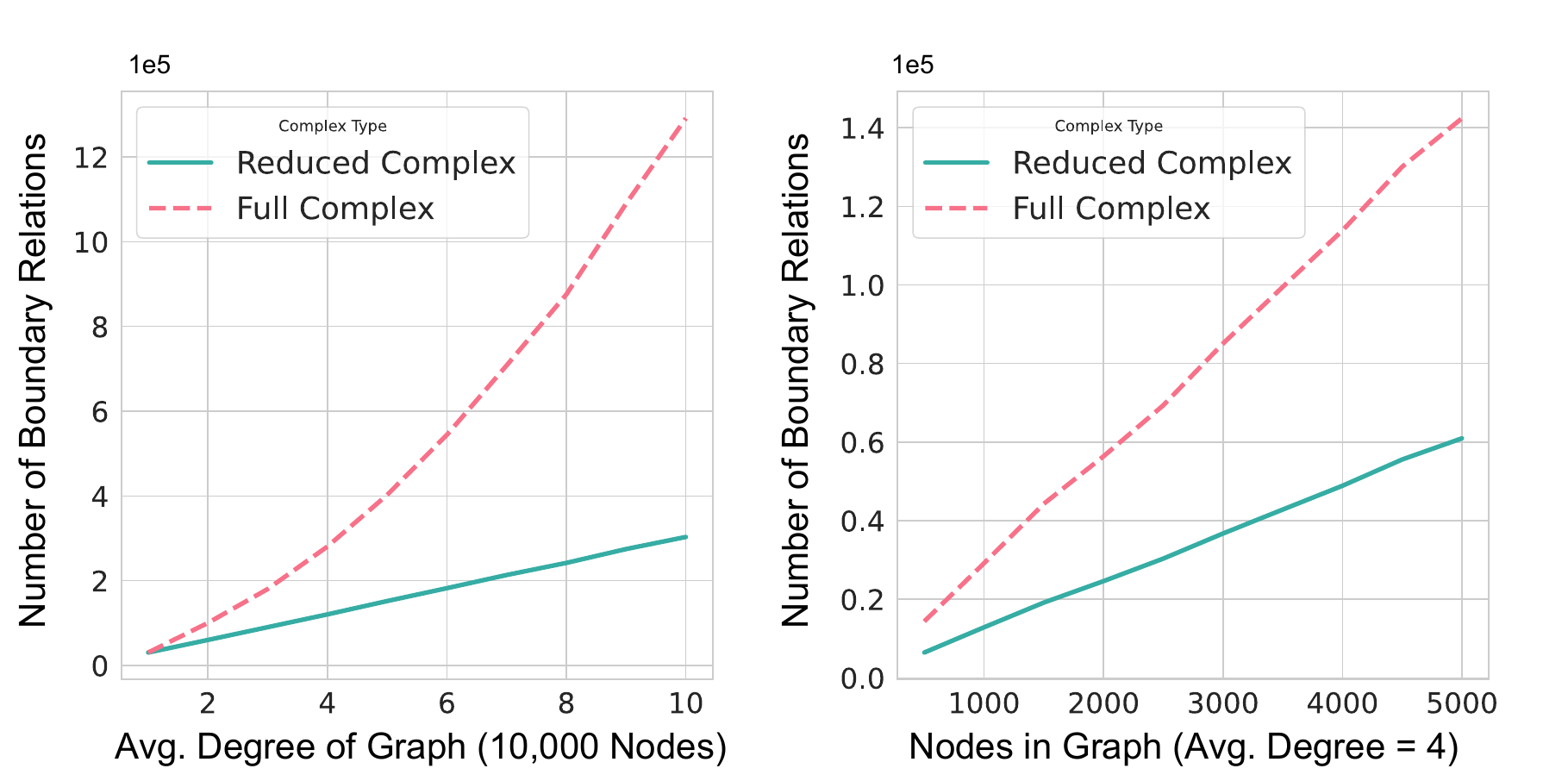}
    \caption{The variation of $|\Sigma|$ for a conventional cell complex and our proposed reduced cell complex with increasing $|E|$ (Theorem \ref{theorem}) (left) and increasing $|V|$ (right), showing that our proposed variation is more space efficient.}
    \label{fig:scaling}
\end{figure}

This operation ensures that $|\Sigma|$ increases \textit{linearly} with $|E|$, making cell complexes much more scalable. We empirically validate these results on large random graphs generated using the Erdős-Rényi method \cite{erdos} and present the results in Figure \ref{fig:scaling}.

Throughout the rest of this text, we use the term \textit{cell complexes} to refer to these restricted/reduced cell complexes, unless stated otherwise. 

\section{Proposed Approach}

\subsection{Lifting Graphs to Cell Complexes}\label{sec:lift}

In this section, we present the lifting algorithm we introduce to create a cell complex $X$ from its underlying graph $G$. Using the lifting algorithm on a given graph $G(V, E)$, we construct $C^{(0)}$, $C^{(1)}$, and $C^{(2)}$ from $V$, $E$, and cycles present in the graph respectively, and add the corresponding $\Sigma$ to create $X(C, \Sigma)$. The lifting algorithm is described in Algorithm \ref{alg:algorithm}. For any given graph, the time complexity of the lifting algorithm is bounded by the time complexity of finding cycles in a graph\footnote{\url{https://networkx.org/documentation/stable/reference/algorithms/generated/networkx.algorithms.cycles.simple_cycles}}. Further details on the algorithm and how cell features are created to perform message passing are deferred to the Appendix.

\begin{algorithm}[h]
\caption{Lifting algorithm}
\label{alg:algorithm}
\textbf{Input}: $G(V, E)$\\
\textbf{Output}: $X(C, \Sigma)$
\begin{algorithmic}[1] 
\STATE $C^{(0)} \gets \phi, C^{(1)} \gets \phi, C^{(2)} \gets \phi, \Sigma \gets \phi$
\FOR{$e_{u,v} \in E$}
\STATE Add $u$, $v$ to $C^{(0)}$
\STATE Add $e_{u,v}$ to $C^{(1)}$
\STATE Add $\sigma_{u,v}$, $\sigma_{u,e_{u,v}}$, $\sigma_{v,e_{u,v}}$ to $\Sigma$
\ENDFOR
\FOR{$c \in$ Set of Cycles in $G$}
\STATE Add $c$ to $C^{(2)}$
    \FOR{$e_{u,v} \in c$}
    \STATE Add $\sigma_{e_{u,v},c}$ to $\Sigma$
    \ENDFOR
\ENDFOR
\STATE $C \gets C^{(0)} \cup C^{(1)} \cup C^{(2)}$
\STATE \textbf{return} $X(C,\Sigma)$
\end{algorithmic}
\end{algorithm}

After lifting the graph to its corresponding cell complex, the next step is to train a \gls{gnn} on cell complexes. Because they are analogous to graphs, we can provide $X$ as an input to any \gls{gnn}, and obtain a trained model $\mathcal{F}$ which will generate representations for cells $C$.

\subsection{Generating Explanations for Cell Complexes}

While graph explainers use diverse methods to generate explanations for \gls{gnn} predictions, they can be abstracted as a function that takes in the trained \gls{gnn} model and graph data and outputs a soft node or edge mask with values between 0 and 1 signifying the importance of each node and edge identified as the explanation. For simplicity, we continue the discussion by defining the explanation as a node mask, an analogous approach can be followed for an edge mask. Formally,
let $\mathcal{M}$ be an explanation mask, then the output of a graph explainer $\mathcal{E}$ can be represented as 

\begin{eqnarray}\label{eq:explainer_eq}
\mathcal{M} = \mathcal{E}(\mathcal{F}, G)
\end{eqnarray}

\begin{figure}[h]
\centering
\includegraphics[width=8cm]{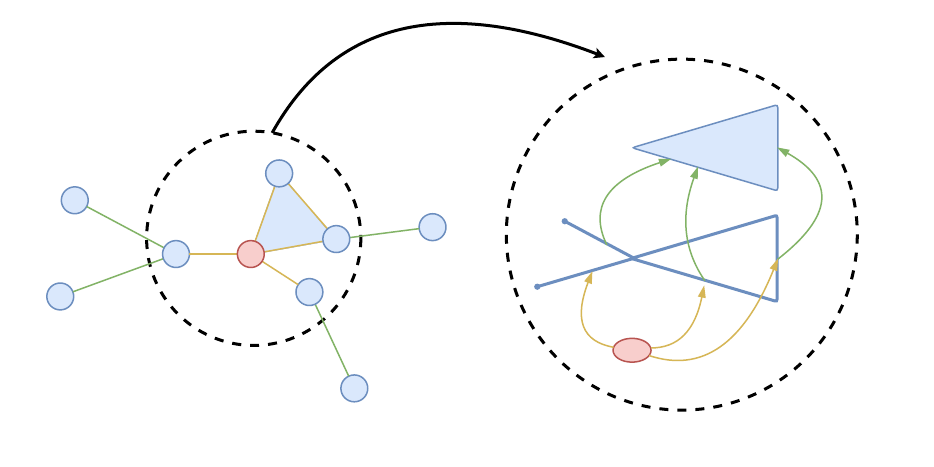}
\caption{Example of a computation cell complex. The figure on the left shows $2$-hop horizontal message passing, while the figure on the right represents $2$-hop vertical message passing, introduced by FORGE. }
\label{fig:computation_graph}
\end{figure}

Graph explainers generate explanations on the \textit{computation graph} of a node, which is the node's $k$-hop neighborhood for a $k$-layer \gls{gnn}. Transitioning from traditional computation graphs, our approach utilizes cell complexes as the input to the function. This shift allows the integration of more complex structural data into the computational framework, through the introduction of \textit{vertical message passing} happening through vertical boundary relations (Section \ref{sec:complex}) in the cell complex, in addition to the already existing (horizontal) message passing, shown in Figure \ref{fig:computation_graph}. Consequently, the learned explanation mask is adapted to represent the computational complex, incorporating both the connectivity and hierarchical structure inherent in cell complexes. Formally, this can be represented by

\begin{eqnarray}\label{eq:explainer_eq_complexes}
\mathcal{M}_X = \mathcal{E}(\mathcal{F}, X)
\end{eqnarray}

By using a cell complex $X(C,\Sigma)$ as input, the explainer outputs an explanation mask $\mathcal{M}_X$, which is an intermediate explanation mask over $C$, indicating which cells are important for the model prediction. 

\subsection{Information Propagation}\label{sec:info_prop}
With the explainer generating an explanation mask $\mathcal{M}_{X}$ for the cell complex $X$, we must propagate this information from the higher-order structures back to the base graph structure to produce the final explanation for $G$. We term this process information propagation, as it involves transferring the learned importance values from the cell complex to the original domain. 

We introduce multiple algorithms to perform information propagation, namely, (1) Hierarchical Propagation, (2) Direct Propagation, (3) Entropy Propagation, and (4) Activation Propagation. For brevity, we expand on (1) in detail and defer the description of other algorithms and optimizations to the Appendix. Hierarchical Propagation is described in Algorithm \ref{alg:hierarchical_prop}.

\begin{algorithm}[h]
\caption{Hierarchical Propagation}
\label{alg:hierarchical_prop}
\textbf{Input}: $\mathcal{M}_X$\\
\textbf{Parameters}: $\alpha_c$, $\alpha_e$\\
\textbf{Output}: $\mathcal{M}$
\begin{algorithmic}[1] 
\FOR{$c^{(2)} \in C^{(2)}$}
    \FORALL{$\sigma_{c^{(1)}, c^{(2)}}$ containing $c^{(2)}$}
    \STATE $\mathcal{M}_X[c^{(1)}] \gets \mathcal{M}_X[c^{(1)}] + \left( \mathcal{M}_X[c^{(2)}] - 0.5 \right) \times \alpha_c$ \label{alg:step1}
    \ENDFOR
\ENDFOR
\FOR{$c^{(1)} \in C^{(1)}$}
    \FORALL{$\sigma_{c^{(0)}, c^{(1)}}$ containing $c^{(1)}$}
    \STATE $\mathcal{M}_X[c^{(0)}] \gets \mathcal{M}_X[c^{(0)}] + \left( \mathcal{M}_X[c^{(1)}] - 0.5 \right) \times \alpha_e$ \label{alg:step2}
    \ENDFOR
\ENDFOR
\STATE $\mathcal{M} \gets \mathcal{M}_X[C^{(0)}]$
\STATE \textbf{return} $\mathcal{M}$
\end{algorithmic}
\end{algorithm}

The main idea behind Hierarchical Propagation is to aggregate the explanations of  2-cells with their 1-cell \textit{boundaries}, then again aggregate explanations of 1-cells with their 0-cell \textit{boundaries}. These 0-cells represent the underlying $0$-skeleton of $X$, which as mentioned in Section \ref{sec:complex} is precisely the set of nodes $V$ of $G(V,E)$. The goal is to utilize cell explanations in a way that polarises important and unimportant nodes in the final explanation mask $\mathcal{M}$. This is done in Algorithm \ref{alg:hierarchical_prop} in steps \ref{alg:step1} and \ref{alg:step2}, where we subtract $0.5$ from the explanation masks of higher order cells in order to quantify how positively or negatively it should influence the overall explanation. They are further multiplied by parameters $\alpha_e$ and $\alpha_c$, which are introduced to have fine-grained control over how much 1-cells and 2-cells, respectively, contribute to the graph explanation. By precomputing the required boundary relations, the time complexity of Information Propagation is reduced to $\mathcal{O}(|E|)$. 

All the steps to obtain explanation $\mathcal{M}$ for a graph $G$ and trained model $\mathcal{F}$ comprising \gls{forge} can be summarised formally using the following equations:

\begin{eqnarray}\label{eq:forge_eq1}
\mathcal{M} = \mathrm{FORGE}(\mathcal{E},\mathcal{F}, G)
\end{eqnarray}

\begin{eqnarray}\label{eq:forge_eq2}
\mathrm{FORGE}(\mathcal{E},\mathcal{F}, G) = \mathrm{PROP}(\mathcal{E}(\mathcal{F}, \mathrm{LIFT}(G)))
\end{eqnarray}

Where $\mathrm{LIFT}$ is Algorithm \ref{alg:algorithm}, $\mathcal{E}$ is any graph explainer, and $\mathrm{PROP}$ is any information propagation algorithm.


\section{Experimental Settings}

\begin{table*}[t]
\centering
\begin{tabular}{cccccccccccc}
\toprule
 & \multicolumn{2}{c}{\textbf{GNNExplainer}} & \multicolumn{2}{c}{\textbf{GraphMask}} & \multicolumn{2}{c}{\textbf{GradExplainer}} & \multicolumn{2}{c}{\textbf{PGMExplainer}} & \multicolumn{2}{c}{\textbf{SubgraphX}} & \textbf{Random}\\
\cmidrule(lr){2-3}\cmidrule(lr){4-5}\cmidrule(lr){6-7}\cmidrule(lr){8-9}\cmidrule(lr){10-11}
\textbf{Datasets} & \textbf{B} & \textbf{F} & \textbf{B} & \textbf{F} & \textbf{B} & \textbf{F} & \textbf{B} & \textbf{F} & \textbf{B} & \textbf{F} &  \\
\midrule
  Benzene & $0.456$ & \underline{$\mathbf{0.772}$} & $0.276$ & \underline{$0.378$} & $0.167$ & \underline{$0.353$} & $0.109$ & \underline{$0.198$} & 0.450 & \underline{$0.614$} & $0.194$ \\
  AlkCarb & $0.054$ & \underline{$0.130$} & $0.048$ & \underline{$0.055$} & $0.114$ & \underline{$0.217$} & $0.095$ & \underline{$\mathbf{0.304}$} & $0.002$ & \underline{$0.011$} & $0.050$ \\
  FluoCarb & $0.207$ & \underline{$\mathbf{0.441}$} & $0.077$ & \underline{$0.230$} & $0.233$ & \underline{$0.439$} & $0.115$ & \underline{$0.300$} & $0.079$ & \underline{$0.095$} & $0.154$ \\
  Mutag  & \underline{$\mathbf{0.380}$} & $0.339$ & $0.127$ & \underline{$0.163$} & $0.172$ & \underline{$0.218$} & $0.114$ & \underline{$0.233$} & $0.079$ & \underline{$0.095$} & $0.112$ \\
\midrule
  Bull/Square & $0.126$ & \underline{$0.433$} & $0.082$ & \underline{$0.241$} & $0.179$ & \underline{$0.298$} & $0.080$ & \underline{$0.150$} & $0.355$ & \underline{$\mathbf{0.447}$} & $0.145$ \\
  House/Hex & $0.107$ & \underline{$0.478$} & $0.115$ & \underline{$0.271$} & $0.199$ & \underline{$0.410$} & $0.088$ & \underline{$0.158$} & $0.346$ & \underline{$\mathbf{0.617}$} & $0.165$ \\
  Wheel/House & $0.102$ & \underline{$0.361$} & $0.233$ & \underline{$0.378$} & $0.169$ & \underline{$0.426$} & $0.083$ & \underline{$0.195$} & $0.246$ & \underline{$\mathbf{0.549}$} & $0.194$ \\
  Cube/Wheel & $0.114$ & \underline{$0.339$} & $0.183$ & \underline{$0.402$} & $0.119$ & \underline{$\mathbf{0.494}$} & $0.091$ & \underline{$0.178$} & $0.397$ & \underline{$0.489$} & $0.221$ \\
\bottomrule
\end{tabular}

\caption{Graph Explanation Accuracy (GEA) ($\uparrow$) scores for baseline explainers \textbf{(B)} against their \gls{forge} variants (\textbf{F)} across all datasets, with \gls{forge} improving performance across various baselines. The best result for each dataset is highlighted in \textbf{bold}. \underline{Underlined} values represent the better result between a base explainer and its \gls{forge} variant.}
\label{tab:results}
\end{table*}

\subsection{Datasets}

\begin{figure}[h]
    \centering
    \includegraphics[width=8cm]{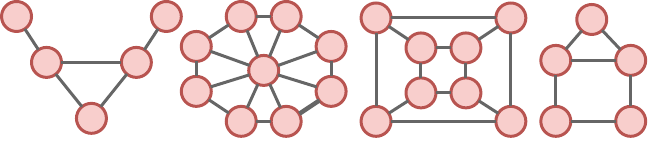}
    \caption{Different motifs used to generate synthetic graphs: (left to right) Bull, Wheel, Cube, and House.}
    \label{fig:motif}
\end{figure}

We evaluate FORGE on both real-world and synthetic datasets for a comprehensive evaluation across diverse conditions. We take real-world datasets from the graph explainability benchmark, GraphXAI \cite{graphxai}, which includes Benzene, Mutagenicity, Alkyl Carbonyl, and Fluoride Carbonyl. For synthetic datasets, we generate random graphs with a distinct subgraph structure, known as a \textit{motif}, which defines the ground truth for the graph. The task involves differentiating between two different motifs. Motifs we test on include Bull, Square (4-Cycle), Hexagon (6-Cycle), Wheel, House, and Cube (Figure \ref{fig:motif}). The synthetic datasets are referred to as Motif1/Motif2 in the text based on the motifs present. Specific details about dataset generation can be found in the Appendix.

\subsection{Evaluation Criteria}

Graph explainer methods produce edge and node masks which represents the most important subgraph that resulted in a model prediction. To evaluate the correctness of this mask, we compare it to the ground truth by adopting two metrics from \citet{graphxai}, Graph Explanation Accuracy (GEA) and Graph Explanation Faithfulness (GEF). GEA uses Jaccard Index \cite{jaccard} to quantify explanation accuracy, and GEF measures how faithful the explanations are to the underlying \gls{gnn} predictor using KL divergence \cite{kldiv}. Further details about the metrics are present in the Appendix.








\subsection{Baselines}

We select a range of established explainer methods as baselines to evaluate FORGE. For perturbation methods, we select GNNExplainer \cite{gnnexplainer}, GraphMask \cite{graph_mask}, and SubgraphX \cite{subgraphx}. For surrogate methods, we evaluate PGMxplainer \cite{pgmexplainer} and for gradient based methods, we select GradExplainer \cite{gradcam}. We use a random explainer as a naive baseline adapted from \citet{graphxai}.

\section{Results}



Table \ref{tab:results} presents the GEA scores for various explainers with and without our framework. The reported results are averaged over $10$ different seeds, with standard deviations reported in the Appendix.
Our experiments reveal several key insights into the performance of different GNN explainers across various datasets.
FORGE consistently improves existing explainers across various datasets with improvements up to $315$\% (FORGE-enhanced GradExplainer on Cube/Wheel), except Mutag for GNNExplainer. The GEF scores are presented in Table \ref{tab:faith}, reinforcing the capabilities of our framework in creating explanations that are comparable or more faithful to the underlying GNN.

\begin{table}[h]
    \centering
    \begin{tabular}{ccc}
    \toprule
       \textbf{Explainer}  & \multicolumn{2}{c}{\textbf{GEF}($\downarrow$)} \\
                    & \textbf{B} & \textbf{F} \\ \midrule
        GNNExplainer & $0.189_{\pm 0.04}$ & $\mathbf{0.083_{\pm 0.02}}$ \\
        GraphMask & $0.051_{\pm 0.03}$ & $\mathbf{0.028_{\pm 0.02}}$ \\
        GradExplainer & $0.389_{\pm 0.13}$ & $\mathbf{0.078_{\pm 0.05}}$\\
        PGMExplainer & $\mathbf{0.204_{\pm 0.05}}$ & $0.234_{\pm 0.03}$ \\
        SubgraphX & $0.008_{\pm 0.01}$ & $\mathbf{0.007_{\pm 0.00}}$ \\
        Random & $0.124_{\pm 0.04}$ & - \\ \bottomrule
    \end{tabular}
    \caption{Graph Explanation Faithfulness (GEF) Scores for the AlkaneCarbonyl Dataset. FORGE-enhanced explainers generate explanations that are comparable or more faithful to the underlying GNN. Values in \textbf{bold} indicate best performance.}
    \label{tab:faith}
\end{table}

Our results reveal substantial variability in explainer effectiveness depending on the graph types and tasks. No single method consistently outperforms others across all datasets, suggesting the importance of choosing explainers tailored to specific problem domains. Perturbation-based methods benefit the most from our framework, with FORGE applied to GNNExplainer generally delivering the best performance on real-world datasets, while FORGE on SubgraphX excels in synthetic datasets. Surrogate and gradient-based methods also show notable improvements across all datasets when enhanced with FORGE. Interestingly, some explainers in their default form perform worse than the Random baseline, a finding supported by the GraphXAI benchmark. However, after applying FORGE, all explainers consistently surpass the Random baseline, achieving significantly improved performance.

For reproducibility of our results, all implementation details are contained in the Appendix. Further experiments and additional results on all datasets for presented experiments are also present in the Appendix.






\section{Ablation Studies}
\subsection{\gls{forge} Components}
\begin{figure}[t]
    \centering
    \includegraphics[width=8.4cm]{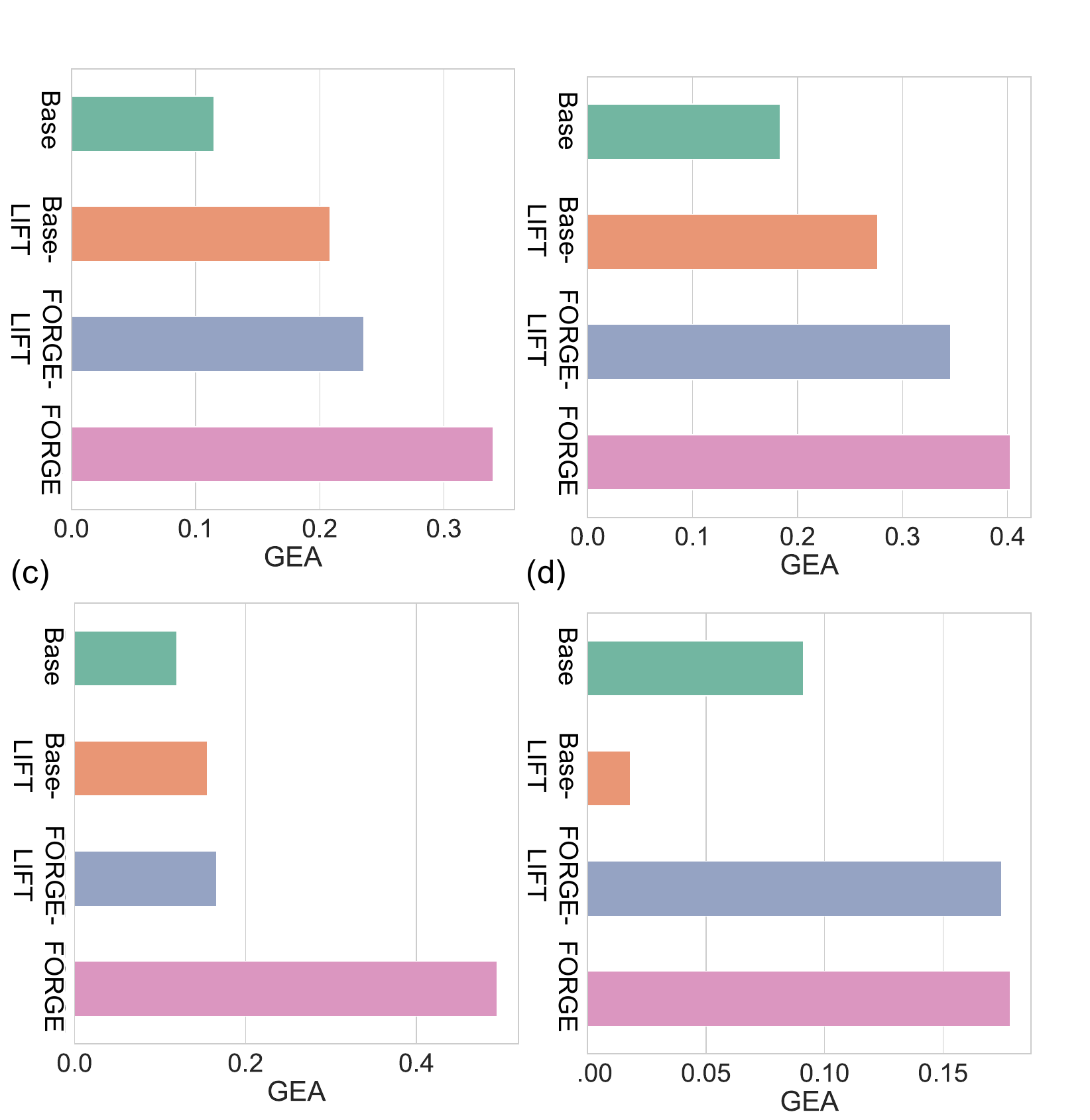}
    \caption{Results of ablations on FORGE Components for \textbf{(a)} GNNExplainer, \textbf{(b)} GraphMaskExplainer, \textbf{(c)} GradExplainer, \textbf{(d)} PGMExplainer on Synth-Wheel/Cube dataset. Both Lifting and Information Propagation contribute significantly to an increase in explanation accuracy. }
    \label{fig:forge_no_prop}
\end{figure}

\gls{forge} comprises two key components: Lifting and Information Propagation. To evaluate their individual contributions, we compare \gls{forge}'s performance against \textbf{Base-LIFT} (a version that applies regular explainers on lifted graphs), \textbf{FORGE-LIFT} (a version of FORGE that only performs lifting), and \textbf{Base} (which has neither Lifting nor Propagation).
Figure \ref{fig:forge_no_prop} demonstrates that FORGE-LIFT, by itself, consistently outperforms the base explainers across all four methods. While Base-LIFT performs better than the baseline, we see that training the underlying GNN on the lifted graphs is important to achieve further performance gains.
This improvement highlights the significant impact of the Lifting component alone in generating more accurate explanations.
Furthermore, when we incorporate Information Propagation to create the full \gls{forge} framework, we observe an additional performance boost on top of Lifting. This enhancement is particularly noticeable in GNNExplainer and GradExplainer (subfigures (a) and (c)).

Interestingly, the impact of Information Propagation appears to vary across different explainers. For instance, its effect seems more pronounced in GNNExplainer and GradExplainer compared to GraphMaskExplainer and PGMExplainer. This variation suggests that the benefits of our proposed propagation strategies may depend on the underlying explanation mechanism.
These findings collectively demonstrate that both Lifting and Information Propagation are crucial components of \gls{forge}, each contributing significantly to the framework's overall performance.
\subsection{Information Propagation Algorithms}
\begin{figure}[t]
    \centering
    \includegraphics[width=8cm]{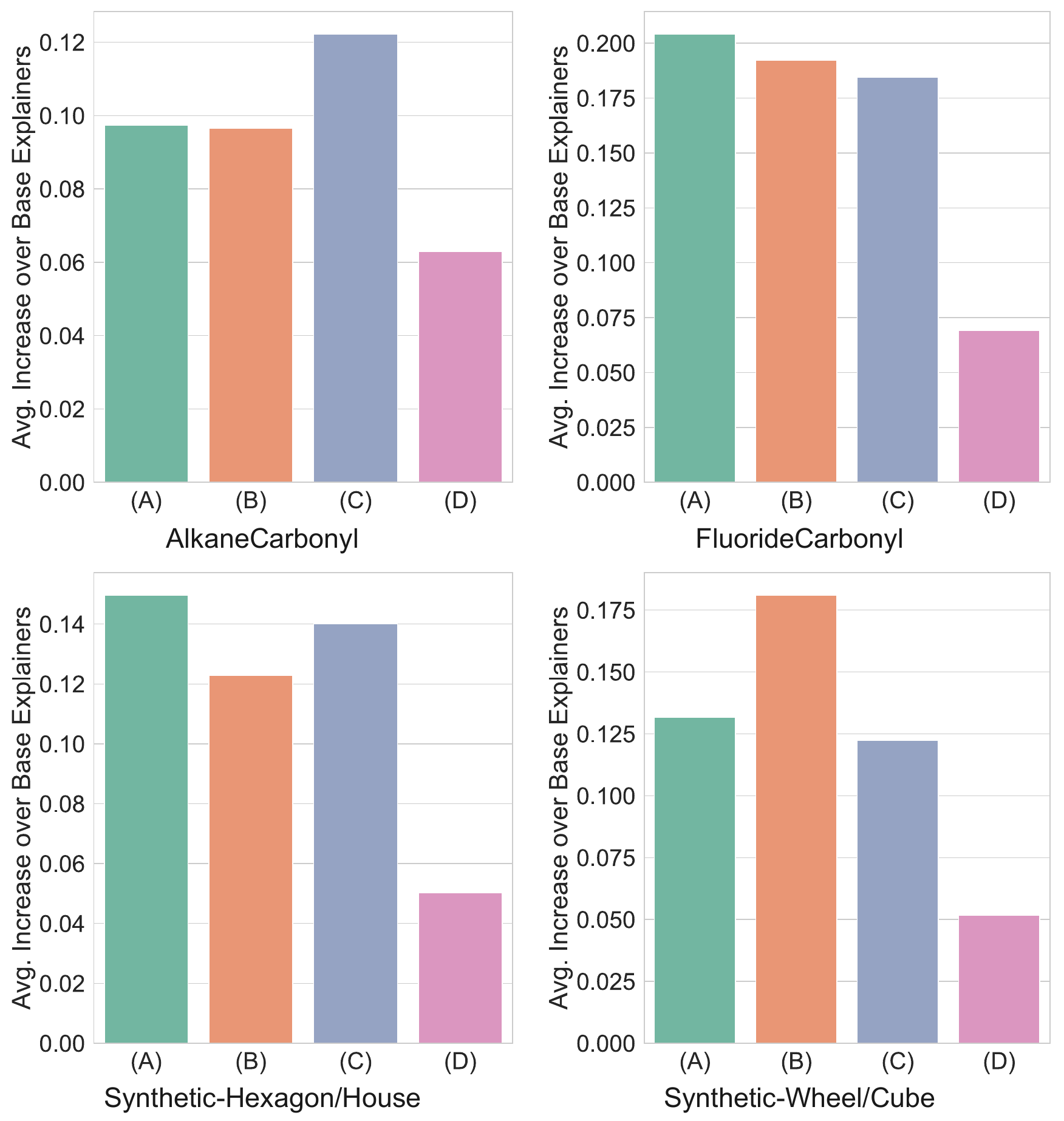}
    \caption{Ablation results for different propagation methods on various datasets, for all baseline explainers. The y-axis represents the average absolute increase in GEA over base explainers. \textbf{(A)} Hierarchical Prop, \textbf{(B)} Direct Prop, \textbf{(C)} Activation Prop, and \textbf{(D)} Entropy Prop.}
    \label{fig:info_prop_ablation}
\end{figure}

Figure \ref{fig:info_prop_ablation} presents our second ablation study, comparing four information propagation algorithms across four different datasets for all explainers, reporting the average GEA. This analysis reveals two key insights: \textbf{(a) Algorithm Performance Variability:} the effectiveness of propagation methods varies significantly across datasets, indicating that no single algorithm consistently outperforms the others in all scenarios, and \textbf{(b) 
 Dataset Dependency:} the optimal choice of propagation algorithm seems to be heavily influenced by the specific dataset being analyzed. This suggests that the graph structure and data properties play an important role in determining the most effective propagation method.

While performance varies, Hierarchical and Direct Propagation methods tend to perform well across most datasets, suggesting they may be more versatile. Activation Propagation shows high effectiveness in certain cases, particularly noteworthy in the AlkaneCarbonyl dataset.

The findings from Figure \ref{fig:info_prop_ablation} emphasise the importance of carefully selecting the propagation method based on the specific dataset and graph structure. They also highlight the need for adaptive or hybrid approaches that can leverage the strengths of different propagation algorithms depending on the context.

%
\section{Conclusion and Future Work}
We introduce FORGE, a novel framework that enhances \gls{gnn} explainability by leveraging cell complexes.
Our framework employs a novel lifting algorithm to convert graphs to cell complexes.
Furthermore, we introduce information propagation algorithms to create more interpretable internal data representations. Our extensive evaluations demonstrate that FORGE consistently enhances explanation accuracy and faithfulness across both real-world and synthetic datasets.
Future work could explore adaptive propagation approaches and investigate FORGE's applicability to diverse graph learning tasks. We hope this work motivates further research into applying higher-order structures for enhanced interpretability.

\section*{Acknowledgments}

We are deeply grateful to the Precog research group for their collective support throughout this work.
We extend special thanks to Shashwat Singh and Vamshi Krishna Bonagiri for their invaluable insights.

\bibliography{aaai25}

\end{document}